\documentclass[preprint]{elsarticle}
\usepackage{subfig}
\usepackage{amsmath}
\usepackage{lineno,hyperref}
\usepackage{booktabs}
\usepackage{pifont}
\usepackage{xcolor}

\newcommand{\cmark}{\ding{51}}%
\newcommand{\xmark}{\ding{55}}%

\modulolinenumbers[5]
\journal{ELSEVIER Signal Processing}

\begin{document}

\begin{frontmatter}

\title{On the Preservation of Spatio-temporal Information in Machine Learning Applications}

\author{Yigit Oktar\fnref{label1}}\ead{Yigit.Oktar@gmail.com}
\fntext[label1]{Department of Computer Engineering, Izmir University of Economics, Izmir, Turkey.}
\author{Mehmet Turkan\fnref{label2}}\ead{Mehmet.Turkan@ieu.edu.tr}\ead[url]{http://people.ieu.edu.tr/en/mehmetturkan}
\fntext[label2]{Department of Electrical and Electronics Engineering, Izmir University of Economics, Izmir, Turkey. Corresponding author.}
%\cortext[cor1]{XXX}
\address{Department of Computer Engineering\\
Department of Electrical and Electronics Engineering\\
Izmir University of Economics, Izmir, Turkey}%\fnref{label3}}

\begin{abstract}
	
	In conventional machine learning applications, each data attribute is assumed to be orthogonal to others. Namely, every pair of dimension is orthogonal to each other and thus there is no distinction of in-between relations of dimensions. However, this is certainly not the case in real world signals which naturally originate from a spatio-temporal configuration. As a result, the conventional vectorization process disrupts all of the spatio-temporal information about the order/place of data whether it be $1$D, $2$D, $3$D, or $4$D. In this paper, the problem of orthogonality is first investigated through conventional $k$-means of images, where images are to be processed as vectors. As a solution, shift-invariant $k$-means is proposed in a novel framework with the help of sparse representations. A generalization of shift-invariant $k$-means, convolutional dictionary learning, is then utilized as an unsupervised feature extraction method for classification. Experiments suggest that Gabor feature extraction as a simulation of shallow convolutional neural networks provides a little better performance compared to convolutional dictionary learning. Many alternatives of convolutional-logic are also discussed for spatio-temporal information preservation, including a spatio-temporal hypercomplex encoding scheme. 
	
  \end{abstract}


\end{frontmatter}

\section{Introduction}

\begin{figure}[!t]
	\centering
	\includegraphics[width=2in]{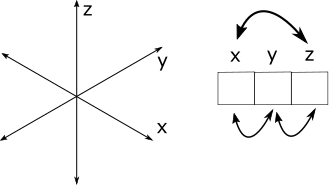}
	\caption{{\footnotesize (Left) There is orthogonal consideration. Every pairwise relation between dimensions is indistinguishable because of orthogonality. (Right) While considering spatial configuration of a $3$-cell $1$D signal, the relation between cells x and z is obviously different from the other relations, i.e., x and z are not neighbors.}}
	\label{fig_spat}
	\end{figure}

	In traditional signal processing and machine learning problems, each data dimension (attribute) is assumed to be orthogonal to others. In other words, there is no distinction between cross-relations of dimensions. While signals carry information through a spatio-temporal configuration, assuming such orthogonality of signal dimensions is highly ill-posed even for $1$D cases. This phenomenon is depicted simply in Fig.~\ref{fig_spat}.
	
	Let us numerically analyze the severity of the problem of casting signals as vectors. Assume that an $n$-sized vector is received through the orthogonality consideration and it is known that the original form is an $n$-sized $1$D signal. If one tries to recover the original spatial configuration without further knowledge (i.e., which value was in which cell), all $n!$ possible spatial configurations are equally likely. This problem becomes even more serious when the dimensionality of the signal itself increases. Consider an $n$-sized vector is received again but the underlying signal is now assumed to be an image. Not only there are permutations involved but also one needs to guess the height and width of the image. In general, for an $n$-sized vector and a $\kappa$-dimensional original signal, the number of possible spatial configurations that the signal could have been in is given as $d_{\kappa}(n)n!$ where $d_{\kappa}(n)$ is the $\kappa$-th Piltz function, which gives the number of ordered factorization of $n$ as a product of $\kappa$ terms~\cite{sandor1996arithmetical}.

	When the above described issue is undertaken, it is not hard to see that many conventional machine learning formulations are highly ill-posed from the perspective of real world signals. Let us now consider the case of $k$-means to be applied on vectorized real world signals, and suppose images for simplicity. As $k$-means originally assumes orthogonality of dimensions, it is easy to apply the usual Euclidean distance metric between vectors. However, it is indeed questionable whether it will capture the notion of distance between two images or rather the average of two images. An example in this light can be given from the domain of Computer Graphics. A direct linear interpolation between two rotation matrices is not natural, thus quaternions are utilized leading to a formulation called spherical linear interpolation~\cite{jafari2014spherical}. A similar consideration might also be superior in the clustering problem of images using $k$-means. However, it is not trivial to cast a general image as a quaternion-like structure for further processing.
	
\begin{figure}[!t]
	\centering
	\subfloat[Main image of $9$]{\fbox{\includegraphics[width=1.2in]{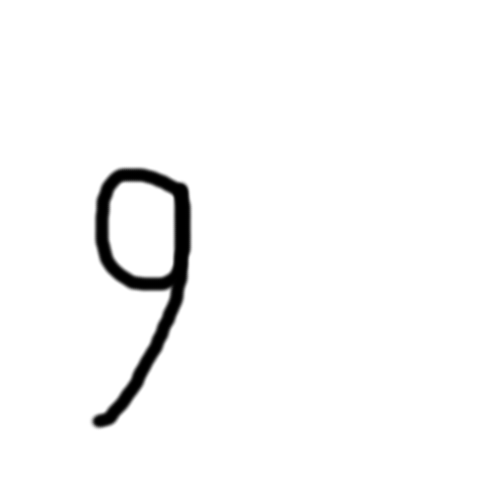}}%
	\label{fig_first_case}}
	\subfloat[The number $8$]{\fbox{\includegraphics[width=1.2in]{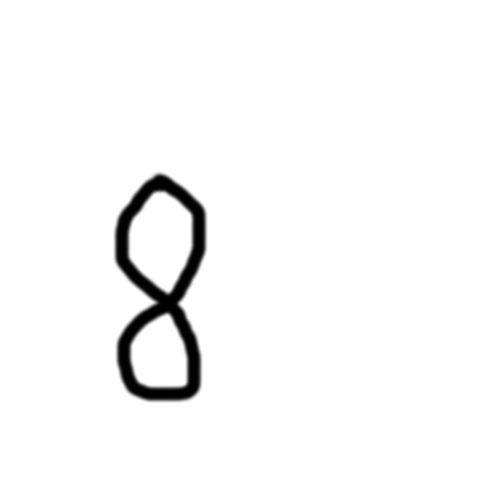}}%
	\label{fig_second_case}}
	\subfloat[Another image of $9$]{\fbox{\includegraphics[width=1.2in]{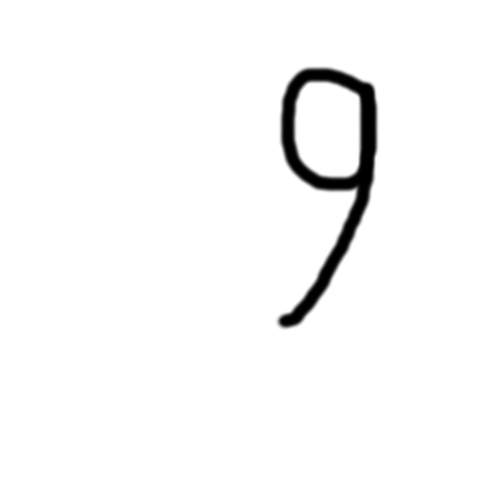}}%
	\label{fig_third_case}}
	\caption{{\footnotesize Vectorized distance will dictate that $8$ is closer to the main image. However, it is indeed more natural to say that two images of $9$ are more similar to each other.}}
	\label{nine_eight}
	\end{figure}

	Let us try to prove that direct vectorized distance calculation is not natural for images by giving a more concrete example. Assume that there is a main image of the number $9$ as exemplified in Fig.~\ref{nine_eight}(a). The question here is which other image is more similar to this main image. Is it the number $8$ in Fig.~\ref{nine_eight}(b) having relatively same spatial position within the frame, or is it the number $9$ in Fig.~\ref{nine_eight}(c) with exact shape but linearly shifted in the frame? Vectorized distance measure will dictate that $8$ is closer to the main image, which is definitely not natural. Therefore, a shift-invariant distance metric could be more powerful in this case.

	For given two images ${\bf I}_a$ and ${\bf I}_b$, the standard (vectorized) Euclidean distance is given in Eqn.~(\ref{Euclid_dist}). This formula can be enhanced with a shift-invariant adaptation as in Eqn.~(\ref{conv_dist}) where ${\bf I}'_a$ denotes the image ${\bf I}_a$ zero-padded on its sides. Alternatively, a shift-invariant distance notion can also be given in terms of inverse of cross-correlation as in Eqn.~(\ref{cross_dist}). Nevertheless, even if a suitable distance metric is found to designate the closest centroid, it is not trivial to obtain the average of a cluster as the new mean for the next step.
	\begin{equation}
	\text{dist}({\bf I}_a,{\bf I}_b)=\|{\bf I}_{a}-{\bf I}_{b}\|_{2} = \sqrt{\sum_{i}{\sum_{j}{({\bf I}_{a}(i,j)-{\bf I}_{b}(i,j))^{2}}}}.
	\label{Euclid_dist}
	\end{equation} 
	\begin{equation}
	\text{dist}({\bf I}_a,{\bf I}_b)=\min_{x,y} \sqrt{\sum_{i}{\sum_{j}}{({\bf I}'_{a}(i+x,j+y)-{\bf I}_{b}(i,j))^{2}}}.
	\label{conv_dist}
	\end{equation}
	\begin{equation}
	\text{dist}({\bf I}_a,{\bf I}_b)=\frac{1} { \max(corr({{\bf I}_a,{\bf I}_b}))}.
	\label{cross_dist}
	\end{equation}
	
	In this study, $k$-means formulation will be considered within a sparse representations framework to provide a self-sufficient shift-invariant version. As noted in earlier studies~\cite{OKTAR201820,oktar2019k,OKTAR2020107634}, the original $k$-means problem can be expressed in a sparse representations framework as a dictionary learning problem. A shift-invariant version of $k$-means can then be derived through a much recent convolutional dictionary learning formulation. It is not a surprise that a convolutional approach leads to a shift-invariant scheme, as convolution is an operator which breaks orthogonality assumption by considering neighboring data points group by group, forming a relation between spatial regions in the signal.
	
	The paper is organized as follows. Section~\ref{sec:CSP} gives the mathematical description of the proposed shift-invariant $k$-means concept, followed by a generalization through convolutional dictionary learning for classification. Section~\ref{sec:ExpRes} details experimental setup and reports experimental results obtained from the proposed concepts. Later, Sec.~\ref{sec:Discuss} discusses many alternatives of convolutional-logic for spatio-temporal information preservation, including a spatio-temporal hypercomplex encoding scheme. Section~\ref{sec:conc} finally concludes this paper with a brief summary.	
	
	\section{Convolutional Sparse Representations}
	\label{sec:CSP}
	
	It is possible to mathematically formulate the conventional $k$-means problem in a sparse representations framework given in Eqn.~\ref{eqn:Kmeans} as follows,
	\begin{equation}
	\begin{gathered}
	{\mathop {\arg\min}\limits_{ {\bf A}, \{{\bf x}_i\}} \sum_i{\| {\bf y}_i-{\bf A}{\bf x}_i \|_2^2 }~~\text{subject to}}\\{\|{\bf x}_i\|_0 = 1~\wedge~\|{\bf x}_i\|_1 = 1~\wedge~{\bf 0} \leq {\bf x}_i,~\forall i},
	\end{gathered}
	\label{eqn:Kmeans}
	\end{equation}
	\noindent where the matrix $\bf A$ is an over-complete dictionary and ${\bf x}_i$ is the sparse representation of the data point ${\bf y}_{i}, \forall i$. Each sparse vector contains only one non-zero component and this component is forced to be positive and sum-to-one. Dictionary columns as atoms (namely ${\bf a}_k$) designate centroids.
	
	While Eqn.~(\ref{eqn:Kmeans}) represents a direct formulation of classical $k$-means, it corresponds to the problematic orthogonality consideration as mentioned previously. A possible shift-invariant alternative of $k$-means is given in Eqn.~\eqref{kmeans_shift_invariant} as follows, 
	\begin{equation}
	\begin{gathered}
	{{\mathop {\arg\min}\limits_{ \{{\bf a}_{k}\}, \{{\bf x}_{i,k}\}} \sum_i{\sum_k{\| {\bf y}_i-{\bf a}_k \star {\bf x}_{i,k} \|_2^2 }}~~\text{subject to}}}\\
	({{k\neq k^* \Rightarrow {\bf x}_{i,k} = {\bf 0})~\wedge~\|{\bf x}_{i,k^*}\|_0 = 1,~\forall i,k},}
	\end{gathered}
	\label{kmeans_shift_invariant}
	\end{equation}
	\noindent where $\star$ denotes the convolution operator and $k^{*}$ is the index of the optimal convolutional atom, or in other words the convolutional centroid that is assigned to the $i^{th}$ data point. Notice here that the non-zero entry of ${\bf x}_{i,k^*}$ is not forced to be $1$, but can now be anything. Therefore, this formulation is not only shift-invariant but also invariant to the magnitude of the pattern. However, this should then be complemented by an atom normalization process.
	
\begin{figure}[!t]
	\centering
	\includegraphics[width=2.5in]{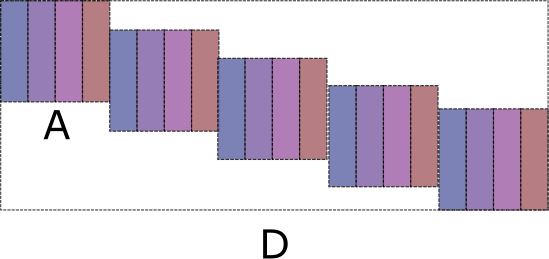}
	\caption{{\footnotesize The local dictionary $\bf A$ consists of convolutional atoms, whereas the global dictionary $\bf D$ is filled with zeros outside the convolutional area.}}
	\label{fig:conv_dict}
	\end{figure}

	Because of the linearity property, atoms in $\bf A$ can also be expressed in a large convolutional dictionary to be denoted by ${\bf D}$ as depicted in Fig.~\ref{fig:conv_dict}. The local dictionary $\bf A$ consists of convolutional atoms, whereas the global dictionary $\bf D$ is filled with zeros outside the convolutional area. In this regard, the mathematical optimization in Eqn.~(\ref{kmeans_shift_invariant}) evolves to Eqn.~(\ref{kmeans_shift_invariant_2}) where $j$ denotes the index of the single non-zero element from the top and $j$ modulo $\#k$ (number of clusters) determines the index of the assigned convolutional centroid $k^*$.
	\begin{equation}
	\begin{gathered}
	{{\mathop {\arg\min}\limits_{ {\bf D}, \{{\bf x}_i\}} \sum_i{\| {\bf y}_i-{\bf D}{\bf x}_i \|_2^2 }~~\text{subject to}}}\\
	{({k^{*} = j \% \#k)~\wedge~\|{\bf x}_i\|_0 = 1,~\forall i}}.
	\end{gathered}
	\label{kmeans_shift_invariant_2}
	\end{equation}

	\subsection{A solution to shift-invariant $k$-means}
	\label{kmeans_si_form}
	
	Since the optimization in Eqn.~(\ref{kmeans_shift_invariant_2}) is highly non-convex, an approximate iterative solution is employed alternating between \textit{assignment to clusters} and \textit{centroid update} akin to Llyod's algorithm for the original $k$-means problem~\cite{jain2010data}. This procedure directly corresponds to \textit{sparse coding} and \textit{dictionary update} steps, respectively, in terms of sparse representations.
	
	In this light, the data assignment step is solved with Orthogonal Matching Pursuit (OMP)~\cite{pati1993orthogonal} assuming $\bf D$ is fixed, to satisfy the $\ell_0$-norm sparsity constraint. On the other side, a straight-forward utilization of conventional dictionary update algorithms, such that Method of Optimal Directions (MOD)~\cite{engan1999method} or KSVD~\cite{aharon2006k}, is not very obvious because the inherent subdictionary $\bf A$ composed of convolutional centroids is only to be updated in ${\bf D}$. To solve this problem, each individual block of the overall sparse representation is extracted as an individual subproblem, on which MOD (i.e., least-squares) update is applied. As the last step, the final updated subdictionary $\bf A$ is attained by averaging all of the resulting individual subdictionaries. To the best of the available knowledge, this naive solution to the centroid update problem is not extensively covered in literature, thus it can be coined as Method of Optimal Subdirections on Average (MOSA).
	
	Experimental results indicate that this adaptation of shift-invariant $k$-means provides better results when compared to its original version for datasets in which considerable shifts exist.

	\subsection{Convolutional dictionary learning as a generalization}
	\label{cdl_form}
	
	Encouraged by the superiority of the shift-invariant $k$-means formulation obtained through a convolutional sparse representation as an unsupervised task, the question is then to generalize this convolutional approach to other machine learning tasks such as classification. The claim is that an unsupervised feature extraction layer that is performed through convolutional dictionary learning as a generalization, can provide superiority over orthogonal-only consideration in also supervised tasks. This claim has already been validated in literature many times~\cite{zeiler2010deconvolutional,pu2016deep,garcia2018convolutional} but an extensive comparison with the classical orthogonality consideration is usually missing.
	
	In this regard, a shift from the strict $\ell_0$-norm constraint to a more lenient $\ell_1$-norm is considered. There are two main reasons behind this decision. First of all, it is unclear how to set the sparsity level in an $\ell_0$-norm formulation since denser choices drastically affect the computational complexity in greedy approaches and sparser solutions can lead to severe information loss. Importantly, most practical studies are based on $\ell_1$-norm in literature~\cite{garcia2018convolutional}.
	
	With the above consideration, a final optimization for convolutional dictionary learning is given in Eqn.~(\ref{conv_bp}) by introducing the $\ell_1$-norm regularization into the formula via a Lagrange multiplier $\lambda$. Iterative solutions which alternates between convolutional sparse coding and dictionary update exist in literature~\cite{garcia2018convolutional}.
	\begin{equation}
	{{\mathop {\arg\min}\limits_{  \{{\bf a}_k\},\{{\bf x}_{i,k}\}}\frac{1}{2}\sum_i{\sum_k{\| {\bf y}_i-{\bf a}_k \star {\bf x}_{i,k} \|_2^2 }}+\lambda \sum_{i,k}{\|{\bf x}_{i,k}\|_{1}}}}.
	\label{conv_bp}
	\end{equation}
	
	In fact, the aim of this study is not to devise new approaches to above optimization but to utilize it as an approach to the orthogonality problem. This unsupervised convolutional decomposition of a signal can be regarded as a feature extraction method that tackles the problem of orthogonality, where the extracted features for the $i^{th}$ data point ${\bf y}_i$ are formed by concatenating the corresponding sparse codes, i.e., ${\bf z}_i = [\{{\bf x}_{i,1}\},\{{\bf x}_{i,2}\}, \dots ]$. Note that concatenation here still assumes orthogonality; however, there now exists a convolutional-logic before the orthogonality consideration which alleviates the main drawbacks of it from the start. The effectiveness of such a layer is to be experimentally tested against various other feature extraction methods in an extensive manner.

	\section{Experimental Results}
	\label{sec:ExpRes}
	
	In the following, two sets of experiments are performed corresponding to the discussions raised in Sec.~\ref{kmeans_si_form} and Sec.~\ref{cdl_form}. All experiments are carried on an Intel(R) Core(TM) i$7$-$6700$HQ CPU @ $2.60$GHz $16$GB RAM machine running on Microsoft Windows $10$ using Matlab $2019$a.
	\begin{figure}[!t]
	\centering
	\includegraphics[width=3in]{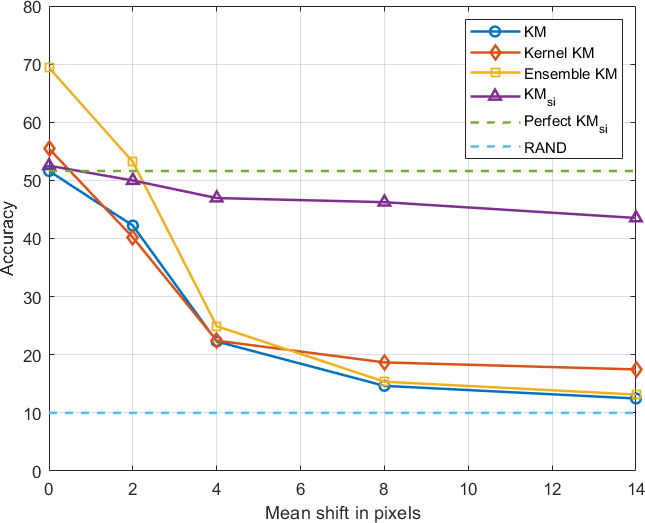}
	\caption{{\footnotesize Clustering accuracy (\%) of $k$-means (KM) based methods as a function of mean shift applied on MNIST. The proposed shift-invariant KM (KM$_{si}$) is robust to shifts.}}
	\label{fig_si_res}
	\end{figure}

	\subsection{Shift-invariant $k$-means}
	\label{results}
	
	In this set of experiments, a dataset is formed by extracting first $1000$ training images of each class from the MNIST handwritten digit database~\cite{MNIST}, making a total of $10000$ images. Four modified versions of this dataset are then obtained to test the shift-invariance property. First of all, empty images of sizes $32\times 32$, $36\times 36$, $44\times 44$ and $56\times 56$ pixels are initialized and original digits are inserted into these widened images with certain uniformly random shifts in \textit{x} and \textit{y} directions. Mean shifts in axes are chosen as $2$, $4$, $8$ and $14$ pixels, respectively, suiting the size of images. The clustering accuracy rates of $k$-means (KM), Kernel KM~\cite{dhillon2004kernel}, Ensemble KM~\cite{iam2010linkclue} and shift-invariant KM (KM$_{si}$) on these cases are illustrated in Fig.~\ref{fig_si_res}.
	
	Not surprisingly, the performance of KM$_{si}$ stays relatively stable in cases of varying shifts, whereas all other methods start to perform poorly when shifts are introduced. It is obvious that a mean shift of $4$ pixels is enough to disrupt the functionality of classical methods for these datasets. Considering original images of sizes $28\times 28$ pixels, this roughly corresponds to a mean shift of $14\%$ of the whole image size. Note also that classical methods perform nearly poor as a random guess method (RAND) in cases of extreme shifts, e.g., $14$ pixels or correspondingly $50\%$. KM has $12.47\%$, Kernel KM has $17.48\%$, Ensemble KM has $13.15\%$ and KM$_{si}$ has $43.53\%$ clustering accuracy in the case of $14$ pixels shift applied on MNIST. This proves that neither kernelization nor ensembles can provide an efficient solution to the shift-invariance problem.
	
	One may argue that a simple preprocessing step, which extracts a precise subimage of the digit in each image, would be enough to sustain shift-invariance for clustering these images; however, such a naive approach cannot be a general solution for natural images. On the other hand, the logic in KM$_{si}$ provides an automatic solution, which is both theoretically and practically sound, without any need for preprocessing. The simplicity and effectiveness of this clustering approach can further pave way to more general techniques with the same logic applied on other machine learning tasks in some settings. In fact, a generalization of KM$_{si}$ via convolutional dictionary learning can be utilized as a powerful unsupervised feature extraction method for classification that alleviates the drawbacks of the classical orthogonality consideration.

	\subsection{Convolutional dictionary learning}
	\label{conv_results}
	
	In this set of experiments, convolutional dictionary learning as an unsupervised feature extraction method is compared against various other well-known feature extraction schemes. An existing library called SPORCO~\cite{wohlberg2017sporco} is utilized for convolutional dictionary learning. In the following reported experiments, linear support vector machine (SVM) classifiers are employed after the feature extraction phase. The motivation behind the linear SVM usage is that, a successful feature extraction must transform the sample space into a linearly separable one as much as possible.
	
\begin{figure}[!t]
	\centering
	\subfloat[Regular - PDL]{\fbox{\includegraphics[height=2.2in]{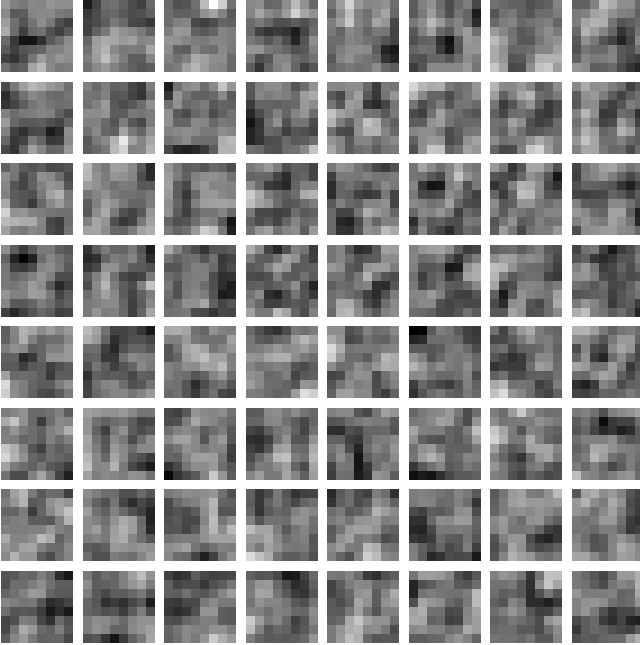}}%
	\label{fig_first_case}}
	\subfloat[Convolutional - CDL]{\fbox{\includegraphics[height=2.2in]{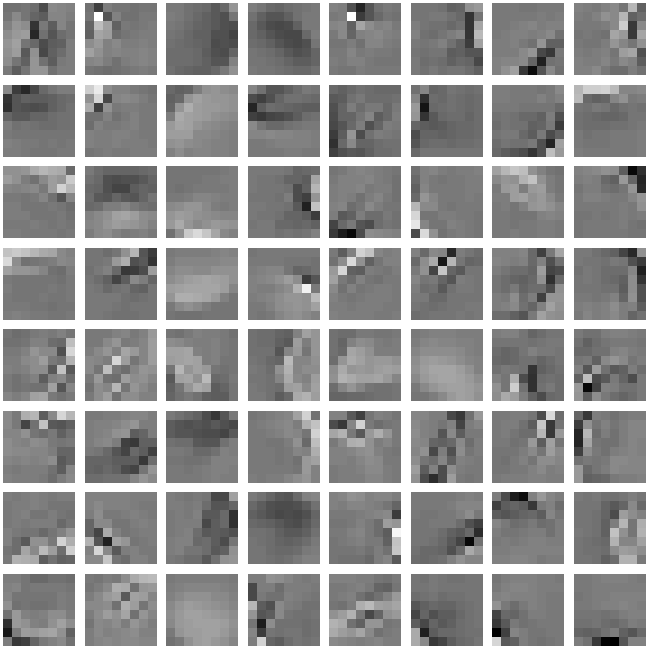}}%
	\label{fig_second_case}}
	\caption{{\footnotesize Patch-based versus convolutional dictionaries learned on MNIST. For a clear visualization, atoms are of size $8\times 8$.}}
	\label{conv_dict2}
	\end{figure}

	There are three employed versions of dictionary learning methods. The global-only dictionary learning (DL) operates over dictionary atoms of size $28\times 28$ pixels, namely atoms cover sample images globally. The patch-based dictionary learning (PDL) trains over dictionary atoms of size $11\times 11$ pixels, where local image patches are extracted in a sliding window manner. This type of approach can be regarded as a local-only one. Both DL and PDL methods are realized through regular dictionary learning iterative steps, i.e., sparse coding and dictionary update. In the proposed method, namely convolutional dictionary learning (CDL), atoms are of size $11\times 11$ pixels but now Eqn.~\ref{conv_bp} is in action instead. While considering the structure of the dictionary in a $2$D form of Fig.~\ref{fig:conv_dict}, CDL can be classified as a both local and global approach. Effects of regular versus convolutional approaches are apparent in the learned atoms at the end of the training process as exemplified in Fig.~\ref{conv_dict2}. Notice that convolutional approach results in filters having Gabor-like appearance.

	Other well-known methods that take spatial information in images into account are Histogram of Oriented Gradients (HOG)~\cite{dalal2005histograms}, Local Binary Patterns (LBP)~\cite{ojala1996comparative} and Gabor Feature Extraction (GFE)~\cite{haghighat2015cloudid}. For HOG, a cell size of $8\times 8$ is chosen with $9$ orientation histogram bins and signed orientation is not used. For LBP, number of neighbors is $8$ and radius of circular pattern to select neighbors is determined as $2$. Rotation information is also encoded. The cell size is $5$ and no normalization is performed. In GFE, a Gabor filter-bank of $15$ filters is employed of size $11\times 11$ with $3$ different scales and $5$ orientations.
	
	Another important categorization of methods is given through whether they perform dimensionality reduction or expansion. The last two methods to be mentioned, namely Autoencoders (AE) and Principal Component Analysis (PCA) both perform dimensionality reduction. Notice that HOG and LBP also accomplish effective dimensionality reduction while other methods instead go through an expansion process. A pooling procedure is closely tied to expansion in case of spatial methods, and is usually performed to reduce the computational cost with the advantage of certain rotation/position invariance. In methods with dimensionality expansion (DL, PDL, CDL, GFE), DL and PDL do not perform an additional pooling since they do not truly preserve spatial configuration. Although PDL takes local spatial information into account, there is no trivial way to perform a meaningful pooling on top. On the other hand, CDL contains a max pooling layer and GFE has an average pooling layer, of cell sizes $2\times 2$ in both cases.
	\begin{table}[!t]
	\caption{{\footnotesize Feature extraction methods in the benchmark.}}
	\footnotesize
	\centering
	\begin{tabular}{@{}lcccccccc@{}}
		\toprule
		& {\bf DL} & {\bf PDL} & {\bf CDL} & {\bf HOG} & {\bf LBP} & {\bf GFE} & {\bf AE} & {\bf PCA}\\ \midrule
		{\bf Learning} & \cmark & \cmark & \cmark & \xmark & \xmark & \xmark & \cmark & \cmark \\
		{\bf Spatial} & \xmark & \cmark & \cmark & \cmark & \cmark & \cmark & \xmark & \xmark \\
		$\#$ {\bf Dimensions} & $5880$ & $5880$ & $2940$ & $144$ & $250$ & $2940$ & $100$ & $100$ \\
		\bottomrule
	\end{tabular}
	\label{conv_dict_info}
	\end{table}

	\begin{figure}[!t]
	\centering
	\includegraphics[width=2.1in]{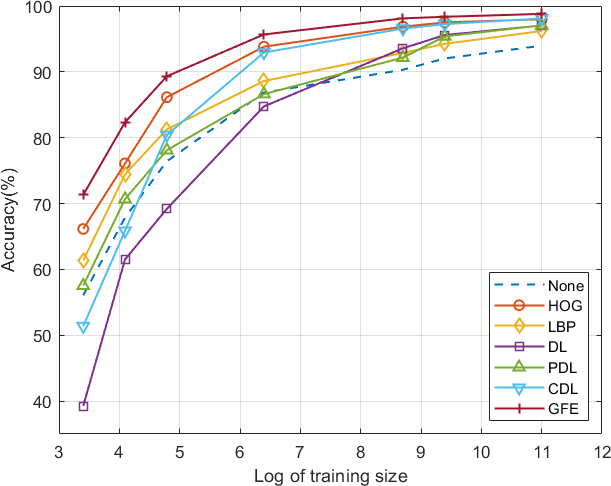}
	\includegraphics[width=2.1in]{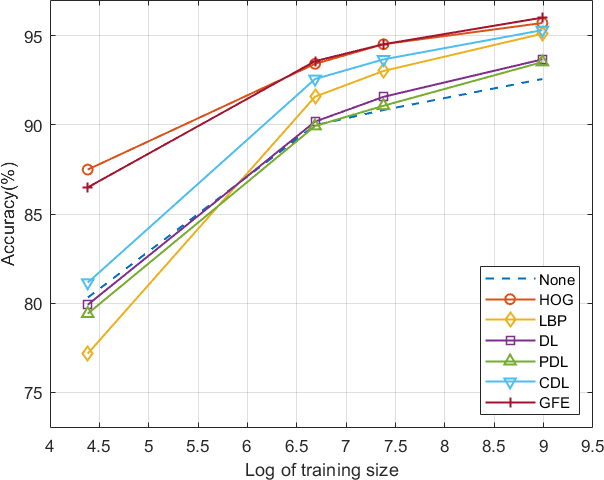}
	\caption{{\footnotesize Classification accuracy ($\%$) as a function of varying training sizes applied on (top) MNIST and (bottom) USPS.}}
	\label{conv_dict_res}
	\end{figure}

	Table~\ref{conv_dict_info} summarizes all feature extraction methods in the benchmark. Note that ``Spatial'' attribute appearing in this table is an antonym for the word ``orthogonality'' in the context of this study. For example, both PDL and CDL can be described as spatial methods since they process images by considering pixels within certain local neighborhoods. However, each pixel is indistinguishable from the others in DL because of the vectorization of the whole frame, resulting in an orthogonality consideration.
	
	After having described all methods in detail, Fig.~\ref{conv_dict_res} depicts classification performance as a function of varying training sizes applied on MNIST and USPS~\cite{USPS} databases. As a global-only dictionary learning method, the inferior performance of DL in case of small training sizes is obvious. A similar behavior is also slightly observable in CDL as a both global and local dictionary learning approach. Although PDL does not perform poorly in small training sizes, it does not provide noticeable advantage over DL in the long run, while CDL outperforms both DL and PDL performing at the capacity of HOG when most of the dataset is used. HOG and GFE together compete for the top performance, whereas CDL performs a little poorer but it is better than LBP. Most importantly, it is apparent that PDL cannot be an alternative to convolutional-logic at least for the $2$D case.
	\begin{table}[!t]
		\caption{{\footnotesize Classification accuracy ($\%$) of feature extraction methods with linear SVM applied on the whole MNIST and USPS datasets.}}
		\footnotesize
		\centering
		\begin{tabular}{@{}lcccccccc@{}}
			\toprule
			{\bf Dataset} & {\bf DL} & {\bf PDL} & {\bf CDL} & {\bf HOG} & {\bf LBP} & {\bf GFE} & {\bf AE} & {\bf PCA}\\ \midrule
			{\bf MNIST} & $97.04\%$ & $97.07\%$ & $98.51\%$ & $97.99\%$ & $96.21\%$ & $98.80\%$ & $94.34\%$ & $94.15\%$ \\
			{\bf USPS}  & $93.67\%$ & $93.52\%$ & $95.31\%$ & $95.71\%$ & $95.11\%$ & $96.01\%$ & $92.02\%$ & $92.72\%$ \\ \bottomrule
		\end{tabular}
		\label{conv_dict_table}
	\end{table}
	
	Table~\ref{conv_dict_table} lists the final classification accuracy results with linear SVM applied on the whole MNIST and USPS databases. GFE is the top performing method as an unsupervised simulation of first layers of a convolutional neural network (CNN). Additionally, CDL and HOG compete for the second place.
	\begin{figure}[!t]
		\centering
		\includegraphics[width=3in]{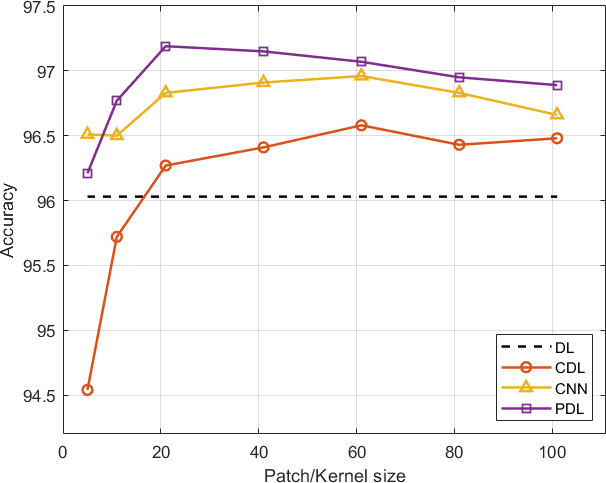}
		\caption{{\footnotesize Classification accuracy ($\%$) as a function of different patch/kernel sizes applied on (preprocessed) MIT-BIH using linear SVM classifiers.}}
		\label{fig_cdl_1d}
	\end{figure}
	
	The convolutional dictionary learning concept is further applied in a $1$D setting. The MIT-BIH arrhythmia dataset~\cite{moody2001impact}, in which the signals correspond to electrocardiogram (ECG) shapes of heartbeats for cases unaffected (normal) and affected by different arrhythmias, is used. These signals are preprocessed and segmented, each segment represents a heartbeat, one of the five different classes~\cite{kachuee2018ecg}.
	
	Preliminary experimentation suggests that the results could be highly dependent on the chosen patch/kernel size as CDL performs poorly for small patch/kernel sizes. These results are summarized in Fig.~\ref{fig_cdl_1d}. In this figure, all methods are devised to be resource-wise equivalent, i.e., they have equal dimensionality of features. DL, PDL and CDL algorithms have the same definitions as in $2$D while they are translated into $1$D equivalent versions. Finally, CNN here denotes a $1$D convolutional neural network as a substructure of a regular $2$D version. For a fair comparison, the architecture of CNN is composed of a convolutional layer, a batch normalization layer, a ReLU layer, a max pooling layer, a fully connected layer, a softmax and a classification layer. In other words, the convolutional-logic is applied once (without getting deep) before the classification stage.
	
	The main observation here is that all spatially-aware methods (PDL, CDL, CNN) outperform the orthogonality consideration of DL, as long as the patch/kernel size is of enough size. It is apparent that a relatively small patch sizes cause CDL to perform very poorly. Such behavior is not observable for CNN which performs well for all kernel sizes chosen. The most surprising result is that PDL outperforms CNN nearly for all cases. However, note that CNN here does not have a deep architecture. The other surprising point is that CDL is the worst among all spatially-aware methods. It is possible that the employed SPORCO library may not be optimized for $1$D settings.
	
\begin{figure}[!t]
	\centering
	\includegraphics[width=3in]{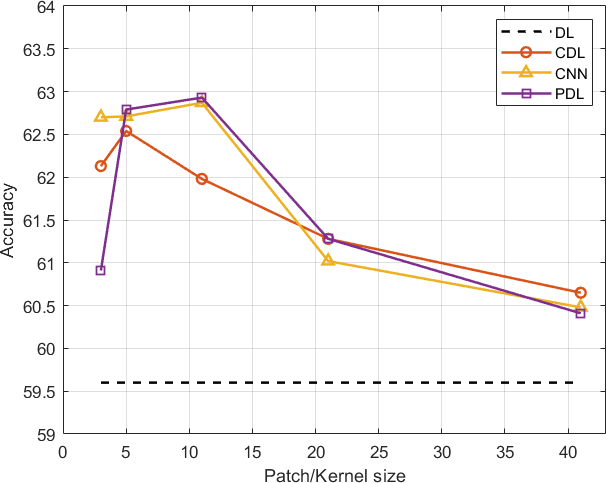}
	\caption{{\footnotesize Classification accuracy ($\%$) as a function of different patch/kernel sizes applied on the raw Electric Devices dataset using linear SVM classifiers.}}
	\label{fig_cdl_1d_2}
	\end{figure}

	To verify the generality of above results, another $1$D problem from a different domain is chosen for the classification of electric devices according to their electric usage profile through raw data. The dataset is obtained from~\cite{UCRArchive} and it contains $8926$ train and $7711$ test samples of size $1\times 96$, with $7$ possible classification labels. In parallel to Fig.~\ref{fig_cdl_1d}, quite similar results are obtained in Fig.~\ref{fig_cdl_1d_2}. With enough patch/kernel size, PDL performance is similar to that of CNN. All methods outperform the baseline of DL.

\begin{figure}[!t]
	\centering
	\includegraphics[width=3in]{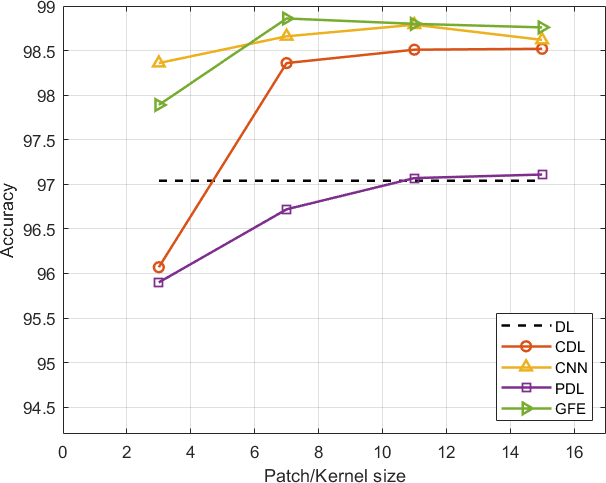}
	\caption{{\footnotesize Classification accuracy ($\%$) as a function of different patch/kernel sizes applied on MNIST using linear SVM classifiers.}}
	\label{fig_cdl_2d}
	\end{figure}

	Inspired by all above experiments measuring the effect of patch/kernel size, the final simulation results on the patch/kernel effect (using the whole MNIST database) are depicted in Fig.~\ref{fig_cdl_2d}. It is clearly observable that CDL nearly matches the performance of a shallow CNN, while PDL performs poorly in this $2$D case. As a conclusion, one can expect PDL as an alternative to CNN in $1$D and CDL in $2$D, as long as patch/kernel size is sensible. Another note is that GFE followed by a linear SVM classifier is a viable unsupervised way of simulating a shallow CNN.

	\section{Discussion on the Spatio-temporal Information Preservation}
	\label{sec:Discuss}
	
	\subsection{Variations on neural networks}
	
	Convolution with a kernel in the input side of a layer corresponds to a locally connected structure instead of a traditional fully connected one. Neighboring cells now occur in a relation, preserving the original spatial configuration. As an alternative to the convolutional approach then, neighboring cells in the input or the output side of a neural network layer can also be put in relation with direct edges in-between, as another way of preserving the original spatial configuration that the input cells have. Possibility of edges in-between in the same layer might force to think of a neural network as a more general directed graph. In fact, this line of logic leads to an alternative structure known as recurrent neural networks (RNN). In most general sense, RNNs represent directed graphs. Note that it is possible to build upon basic RNN structure through bidirectional logic~\cite{schuster1997bidirectional} and long-short term memory concept~\cite{sak2014long}.
	
	On the other hand, empirical evaluation suggests that temporal convolution, or in other words $1$D convolutional-logic surpasses the capacity of recurrent architectures in sequence modeling~\cite{bai2018empirical}. It is still an open question whether temporal dimension should be regarded as just another spatial dimension or whether a hybrid approach is better. This is rather a deep issue related to properties of space and time. Instead, considering neural networks of any structure as directed and possibly cyclic graphs, or in other words as neural graphs, might pave way to better understanding of the brain. Note that this concept is rather different than graph neural networks which use graphs as inputs~\cite{scarselli2008graph}.
	
\begin{figure}[!t]
	\centering
	\includegraphics[width=3in]{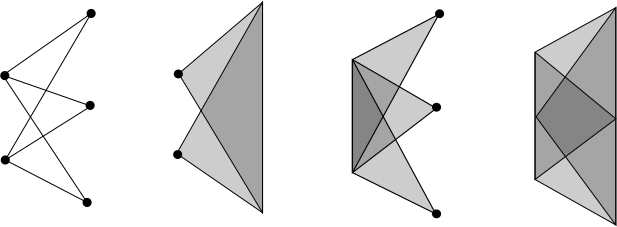}
	\caption{{\footnotesize A generalization of neural network layer cases. (From left-to-right) Discrete-discrete (classical), discrete-continuous, continuous-discrete and continuous-continuous input and output layers.}}
	\label{fig_cases}
	\end{figure}

	Another generalization for neural networks is possible by considering infinite width neural networks~\cite{arora2019harnessing}. Recent results suggest that deep neural networks that are allowed to become infinitely wide converge to models called Gaussian processes~\cite{lee2017deep}. However, such studies do not consider the case when there are in-between connections within layers. Considering the existence of these connections, this can further lead to having an infinite but continuous (input or output) layers, which is indeed applicable mathematically and practically. A generalization of neural network layer cases in this sense is depicted in Fig.~\ref{fig_cases}. The third case in this figure is important in that, it leads to the concept of functional machine learning. This alone may not be enough to preserve the spatial configuration of the input layer. Therefore, additional locally connected versions of these structures can also be proposed.

	\subsection{Multilinear approach}
	
	\subsubsection{Tensor-based sparse representations}
	
	The fact is that images are not vectors, thus vectorization breaks the spatial coherency of images which is investigated by~\cite{hazan2005sparse}. This line of thought is centralized around tensor factorization as a generalization. The study in~\cite{hazan2005sparse} reports that by treating training images as a $3$D cube and performing a non-negative tensor factorization (NTF); higher efficiency, discrimination and representation power can be achieved when compared to non-negative matrix factorization (NMF).
	
	There are two main branches of tensor decomposition. In the first branch, studies are based on canonical polyadic decomposition (CPD), sometimes also referred to as CANDECOMP/PARAFAC~\cite{kolda2009tensor}. The most relevant example from literature is K-CPD~\cite{duan2012k}, an algorithm of overcomplete dictionary learning for tensor sparse coding based on a multilinear version of OMP and CANDECOMP/PARAFAC decomposition. K-CPD surpasses conventional methods in a series of image denoising experiments. Most recently, a similar framework is also successfully utilized in tensor-based sparse representations for classification of multiphase medical images~\cite{wang2020tensor}. The second branch is centered around the Tucker decomposition model instead, which is a more general model than CPD~\cite{caiafa2013computing}. The study in~\cite{caiafa2012block} presents the foundations of the Tucker decomposition model by defining the Tensor-OMP algorithm which computes a block-sparse representation of a tensor with respect to a Kronecker basis. In~\cite{caiafa2013computing}, authors report that a block-sparse structure imposed on a core tensor through subtensors provide significant results. The Tucker model together with block-sparsity restriction may work significantly well, since the higher dimensional block structure is meaningfully applied on the original sparse tensor in the form of subtensors. There are many other studies in literature specifically based on the Tucker model of sparse representations with or without block-sparsity and additionally including dictionary learning~\cite{qi2013two,roemer2014tensor,peng2014decomposable}.
	
	Certain parallels can be drawn between convolutional dictionary learning and tensor-based sparse representations. As an example, the study in~\cite{huang2015convolutional} proposes a novel framework for learning convolutional models through tensor decomposition and shows that cumulant tensors have a CPD whose components correspond to convolutional filters and their circulant shifts.
	
	On the other side, tensor-based approaches (both CPD and Tucker models) do not still provide a solution to $1$D case. Without loss of generality, let us assume that the signal is in the form of a column vector ${\bf s}$. Since the signal is one-dimensional, there will be a single matrix $\bf D$ for that single dimension in the Tucker model. Therefore, the model attained is ${\bf s} = {\bf x} \times_{1}{\bf D}$ in Eqn.(\ref{tensor_fin}). It is also possible to show that ${\bf x}\times_{1}{\bf D} = {\bf D}{\bf x}$. From the CPD model perspective, there is equivalently $\sum_{i}{x_{i} {\bf d}_{i}^{(1)}}$ where $x_{i}$ is the single sparse coefficient associated with $i^{th}$ atom ${\bf d}_i$. Hence, one arrives at a standard formulation in Eqn.(\ref{tensor_fin}), namely Tucker and CPD models are equivalent in one-dimensional case, all corresponding to conventional orthogonal sparse representation.
	\begin{equation}
	{\bf s} = {\bf x}\times_{1}{\bf D} = {\bf D}{\bf x} = {\sum_{i}{x_{i} {\bf d}_{i}^{(1)}}}
	\label{tensor_fin}
	\end{equation}
	
\begin{figure}[!t]
	\centering
	\includegraphics[width=4in]{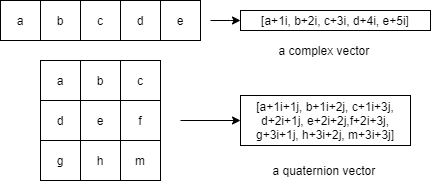}
	\caption{{\footnotesize An encoding scheme to preserve spatio-temporal information for (top) $1$D mono audio and (bottom) $2$D grayscale image cases.}}
	\label{hyper_enc}
	\end{figure}

	The above observation brings up an important question onto the table. Although tensor-based approaches provide advantage when the signals are multidimensional, these formulations will not provide an edge for $1$D signals. The remedy may come from considering a $1$D signal, not as a $1$D vector of elements solely. In other words, a $1$D complex vector can be formed by coding the cell positions in the imaginary parts to overcome the orthogonality problem in standard $1$D vector representation as depicted in Fig.~\ref{hyper_enc}. This paves way to performing sparse representations of complex valued data, or even quaternion valued data, to accommodate more information in cases of higher dimensionality. Utmost generalization is achieved through geometric algebra as a generalization of hypercomplex numbers.

	\subsubsection{Complex, hypercomplex and geometric algebra based approaches}
	
	Note that quaternion algebra is the first hypercomplex number system to be devised that is similar to real and complex number systems~\cite{moxey2003hypercomplex}. The study in~\cite{xu2015vector} states that a quaternion-based model can achieve more structured representation when compared to a tensor-based model. Comparisons between quaternion-SVD and tensor-SVD~\cite{kilmer2011factorization} provide their equivalence, but superiority of quaternion-SVD arises when it is combined with the sparse representation model. It is possible to formulate a quaternion-valued sparse representation of color images that surpasses the conventional logic~\cite{xu2015vector}.
	
	There are four possible models to represent color images as suggested in~\cite{xu2015vector}. The first one is the monochromatic model, in which each color channel is represented separately. The second one is the concatenation model, where a single vector is formed by concatenating three color channels~\cite{mairal2007sparse}. The third is the tensor-based model, where the color image is thought of as a $3$D cube of values. The last one is the quaternion-based model, where each color channel is assigned to each imaginary value, i.e., r,g,b to i,j,k respectively. Most importantly, all these models are analytically unified.
	
	There is also one more possible model that is subtler. As depicted in Fig.~\ref{hyper_enc}, one can encode a mono audio as a vector of complex numbers where imaginary values indicate the timed position, in a similar way one can encode a grayscale image as a quaternion-valued vector where imaginary parts are allocated to indicate the pixel positions. While thinking of a color image as a $3$D cube, there is a possible quaternion-based model in which imaginary units encode the position within this cube and the scalar denotes the value of that cell. The same quaternion-based encoding can be applied to any $3$D scalar data.
	
	For further machine learning in this proposed scheme, a hypercomplex to real feature extraction layer is required since current mainstream classification algorithms need real-valued data. Another option is to consult classification algorithms that can directly handle hypercomplex values. This line of logic paves way to consider complex/hypercomplex valued neural networks as viable tools~\cite{hirose2012complex,isokawa2003quaternion}. As a future work, comparison of spatio-temporally encoded hypercomplex neural networks with conventional convolutional or recurrent neural networks may lead to deeper understanding of the deep learning concept. As a motivation, a single complex-valued neuron can solve the XOR problem~\cite{nitta2003solving}. In addition, the fact that quaternions can be used to implement associative memory in neural networks is promising~\cite{chen2017design}.
	
	Another line of generalization can deal with the case when the data has more than three dimensions. In such a case, a quaternion is not enough to designate the cell position and its value. As an extension, octonion algebra can accommodate up to seven imaginary channels~\cite{popa2016octonion,lazendic2018octonion}; however, loses the associativity property. The study in~\cite{lazendic2018hypercomplex} reports that all algebras of dimension larger than eight lose important properties, since they contain algebras of smaller dimension as subalgebras. This might be an issue related to physics of space and time, which is out of scope of this study. The important fact is that the domain dealing with generalization of hypercomplex numbers is called ``geometric algebra'' and is gaining great attention lately~\cite{wang2019geometric}.

	\section{Conclusion}
	\label{sec:conc}
	
	This study aims to draw attention to orthogonal viewpoint that is taken by many machine learning methods, such as $k$-means. Convolution operator can be used as a remedy for this problem, as it partially preserves the spatio-temporal information inherent in signals. However, one may need to find alternatives to convolutional approaches in order to further increase the understanding on this subject. Spatially sparse connections in neural networks might be an alternative. A continuous to discrete generalization of a neural network layer can also pave way to the concept of functional machine learning. Most importantly, analytic approaches such as multilinear formulations must be thoroughly investigated as alternatives. In fact, to compare methods assuming orthogonality with convolutional-logic, first of all hypercomplex versions of classical methods must be considered where imaginary parts of hypercomplex values encode the spatio-temporal placement. As noted before, $1$D case might be a crucial case not to be underestimated.
	
	Going back to the clustering problem, one should now notice that shift invariant $k$-means can include rotation invariance as a more general formulation~\cite{barthelemy2012shift}. Interestingly, the study in~\cite{bar2010hierarchical} notes that a log-polar mapping converts rotations and scalings to shifts in \textit{x} and \textit{y} axes respectively; therefore, invariance under general transformations is possible. In the bigger picture, convolutional-logic or other frameworks that sustain invariance is related to two-stream hypothesis (i.e., where pathway and what pathway), a model of the neural processing of vision as well as hearing~\cite{eysenck2005cognitive}. In other words, a spatio-temporal information preserving perspective on the clustering problem brings us closer to inner working principles of the brain. Also related to convolution, $n$-dimensional generalization of Gabor filters can be investigated as a future work.
	
	A final general note is the distinction between analysis versus synthesis sparse models. Throughout this study, the synthesis model is used of the form ${\bf Y} = {\bf A}{\bf X}$ where ${\bf X}$ is sparse. However, there is also the analysis model having the form ${\bf A}{\bf Y} = {\bf X}$, in which the dictionary ${\bf A}$ multiplied by the input ${\bf Y}$ now results in the sparse codes in ${\bf X}$~\cite{shekhar2014analysis,gu2017joint}. Such model is closer to neural network formulations, and further investigation of analysis model might pave way to a unified perspective on sparse representation models which also includes neural networks.

\footnotesize
\bibliography{mybibfile}

\end{document}